\documentclass[journal]{IEEEtran}
\usepackage[T1]{fontenc}
\usepackage{amsmath,amssymb,amsfonts}
\usepackage{bm}
\usepackage{graphicx}
\usepackage{booktabs}
\usepackage{multirow}
\usepackage{xcolor}
\usepackage{algorithm}
\usepackage{algpseudocode}
\usepackage{tikz}
\usetikzlibrary{arrows.meta,positioning,calc,decorations.pathmorphing,shapes.geometric,fit,backgrounds}
\usepackage{cite}
\usepackage[hidelinks]{hyperref}

\definecolor{mnavy}{HTML}{25548F}
\definecolor{mgray}{HTML}{BDBDBD}
\definecolor{mjade}{HTML}{2AA198}
\definecolor{mochre}{HTML}{CC8A00}
\definecolor{mburg}{HTML}{9E2A3F}
\colorlet{mteach}{black!58}

\begin{document}
\bstctlcite{IEEEexample:BSTcontrol}

\title{Model Annealing: Data-Efficient Learning through Review and Consolidation}

\author{Ali {Mohammadi Ruzbahani} and Rezvan Abouzadeh%
\thanks{A. Mohammadi Ruzbahani is with the Smart Cyber-Physical Systems (SCPS) Lab, University of Calgary, Calgary, AB, Canada, and is also the Chief Technology Officer of MetaKet Inc., Calgary, AB, Canada (e-mail: ali.mohammadiruzbaha@ucalgary.ca).}%
\thanks{R. Abouzadeh is a Senior Engineer with MetaKet Inc., Calgary, AB, Canada (e-mail: rezvan@metaket.com).}}

\maketitle

\begin{abstract}
A small labelled dataset is memorized long before the training budget runs out, and the one-hot labels that remain to be fitted then mostly sharpen confidence. Model annealing spends the rest of the budget on review and consolidation, relearning from the network's own knowledge. Training is divided into episodes whose learning rate is raised again and cooled, every cooled state joins a memory of past selves, and each new episode learns from the labels, from the tempered predictions of this memory, and from adversarial dreams sought where student and memory disagree most. The memory averages predictions rather than weights, and the deployed network is the last student, so inference costs nothing extra. Across CIFAR-100, Tiny-ImageNet and Imagewoof, four architecture families and 5\% to 100\% of the labels, model annealing raises the top-1 accuracy of the AdamW-trained ConvNeXt, ViT and MLP-Mixer over cross-entropy training in 28 of 33 configurations, by 1.3 points on average, and is on par in accuracy with snapshot distillation, the closest competitor. On these families it is better than both in likelihood, calibration and robustness to corruptions and small adversarial perturbations. For an SGD-trained ResNet-18 it lowers accuracy but still improves robustness to small adversarial perturbations. Separating the reheated schedule from relearning shows that relearning helps every family and that the losses come from reheats, which undo most of this network's fit to its training set. How early the accuracy of a cross-entropy run saturates largely separates the AdamW families from ResNet-18 but does not predict the gain within them. With little data, relearning from the latest snapshot without dreams, as snapshot distillation does, compounds overconfidence in the AdamW families from episode to episode. The memory and the dreams each curb most of this drift, and together they remove it.
\end{abstract}


\section{Introduction}

\IEEEPARstart{T}{his} paper trains deep networks on small labelled datasets in a cycle borrowed from human study: read the material, review it and work problems on it, let it settle, and return to it to build new knowledge on what was retained. Evidence on how people learn favors working with material over rereading it. An assessment of study techniques rates self-testing, such as answering practice questions on the material, among the most useful and rereading among the least \cite{dunlosky2013improving}. In one experiment, restudying foreign words that students could already recall did not improve their recall a week later, whereas recalling the words again improved it greatly \cite{karpicke2008critical}. In another, students who reread a prose passage were more confident of remembering it than students tested on it, yet a week later they recalled less \cite{roediger2006test}. New memories, moreover, stabilize only gradually after learning, a process known as consolidation \cite{mcgaugh2000memory}. The cycle serves here as an organizing analogy, not as a model of human memory.

Plain supervised training does only the reading, over and over. A deep network often memorizes a small labelled dataset long before its training budget is spent. From then on the one-hot labels tell it little that is new, and the remaining epochs mostly sharpen its confidence. Like the students who reread, the network grows more confident without growing more accurate, so small-data training can fail twice: accuracy plateaus early, and the predictions become overconfident. The regime is common wherever labels are expensive or cannot be pooled. Runtime anomaly detectors for cyber-physical systems are typically learned from normal operating data alone \cite{ruzbahani2026pia}, and organizations that would gain from each other's data are often unwilling to share raw data, gradients or model parameters at all \cite{ruzbahani2026dtqcl}.

Model annealing trains a network through the whole cycle. Its budget is divided into equal episodes, and the first episode, ordinary supervised training on the labels, is the first reading. Each later episode is a new study session in which the network, now a student of its own past, starts with a reheat: the learning rate returns to its peak, which frees the network to leave the solution it settled into. The session is spent on relearning, which combines review and practice. To review, the student learns not only from the labels but also from the tempered predictions of a memory of every state it has cooled into, an ensemble of past selves whose older members fade geometrically but never vanish. Review differs from rereading in what it presents: a tempered distribution does not saturate the way a one-hot label does, and it still says how the classes relate, so review gives the student something to learn after the labels are memorized. To practice, the student works problems called dreams. A one-step adversary searches a small neighborhood of a training image for the input on which the student departs most from its memory, and the memory labels that input, so dreams are problems off the training set, sought where current and past knowledge disagree most.

While the student reviews and practices, the learning rate cools along a cosine to nearly zero and the network settles. Every episode, the first included, ends in consolidation: a frozen copy of its cooled state joins the memory that all later episodes review. The training data never change, so the new knowledge an episode adds is what the network learns in it beyond the earlier episodes; consolidation keeps it for every later review. The deployed model is the last student, so inference costs no more than for a single network.

Training a network to match its own earlier predictions has a history. Born-again networks distill a model into an identical copy over several generations \cite{furlanello2018born}, snapshot distillation and progressive self-knowledge distillation reuse earlier states of the same run as teachers \cite{yang2019snapshot,kim2021pskd}, and knowledge evolution re-initializes part of a network over many generations to help small datasets \cite{taha2021ke}. In image classification these methods were evaluated mostly on convolutional networks, and two questions that decide whether self-relearning is worth its cost remain open: which of its ingredients are responsible for the gains, and for which architectures and amounts of data the gains turn into losses. Model annealing gathers the ingredients into one framework in which each can be switched on and off, and we use it to answer both questions.

Our contributions are the following.
\begin{itemize}
\item Model annealing, a single-run procedure that reheats and cools the learning rate in equal episodes, consolidates every cooled state as a frozen member of a function-space memory of past selves, and relearns from this memory and from adversarial dreams, at the inference cost of one network.
\item A controlled study over three benchmarks, four architecture families and four amounts of labelled data, with paired tests, in which model annealing improves accuracy and likelihood over cross-entropy training for the three AdamW-trained families, where it is on par with snapshot distillation in accuracy but better in likelihood, calibration and robustness, and improves robustness to small adversarial perturbations for all four.
\item A decomposition into a schedule effect and a relearning effect, showing that relearning helps every family and that the losses on ResNet-18 come from reheats that undo most of its fit, together with evidence that early saturation of a plain run largely separates the AdamW-trained families, where model annealing pays, from ResNet-18 but does not predict the gain within them.
\item Evidence that, for the AdamW-trained families with little data, relearning from the latest snapshot without dreams compounds overconfidence over episodes, that a memory of all past selves and the dreams each curb most of this drift and together remove it, and that the memory opposes the labels late in each episode, where it restrains memorization.
\end{itemize}

\section{Related Work}

Knowledge distillation trains a student to match the temperature-softened output distribution of a teacher \cite{hinton2015distilling}. Born-again networks showed that a student with exactly the teacher's architecture can outperform it, and that the procedure can be repeated over several generations \cite{furlanello2018born}. Each generation, however, costs a full training run. Snapshot distillation avoids that cost by dividing one run into cycles and letting the network at the end of a cycle teach the next one \cite{yang2019snapshot}, while progressive self-knowledge distillation softens the hard targets with the model's own predictions from the previous epoch, weighting them more as training proceeds \cite{kim2021pskd}. Predictions can also replace parameters as the carrier of knowledge between models, as in query-based collaborative learning among mutually distrustful parties \cite{ruzbahani2026dtqcl}. Model annealing belongs to this family but changes what the student relearns from. Its teacher is a memory of all past selves rather than the latest snapshot, and part of the supervision is spent on inputs chosen adversarially instead of on the training images alone.

A second line of work restarts or perturbs optimization deliberately. Warm restarts of the learning rate improve the anytime performance of stochastic gradient descent \cite{loshchilov2017sgdr}. Shrinking the weights and adding noise before training resumes closes the generalization gap that warm-started networks otherwise suffer \cite{ash2020warm}, and knowledge evolution splits a network into a retained part and a re-initialized part, retraining it over many generations with clear gains on small datasets \cite{taha2021ke}. The name of our method comes from simulated annealing \cite{kirkpatrick1983sa}, where controlled heating followed by slow cooling lets a search leave poor local minima before it settles.

How the past states of a run should be combined is usually answered in weight space. Reptile moves an initialization toward the weights reached after a short inner training loop \cite{nichol2018reptile}, and averaging the weights visited under a cyclical or high constant learning rate yields flatter solutions that generalize better \cite{izmailov2018swa}. Language models have recently been placed in the role of the meta-learner, writing training-data-free weight updates from the snapshots of a centralized run \cite{ruzbahani2026gradreason} and aggregation rules over client models in federated learning \cite{ruzbahani2026fedreason}; in the centralized case, the update the model arrived at in every run was an average of recent weight snapshots. Such averages are safe only between networks joined by a path of low loss. Copies of a network trained from one state under different mini-batch noise share a linear path of this kind only when that state lies sufficiently far into training, and copies trained from initialization do not \cite{frankle2020linear}. Snapshot ensembles avoid the question by averaging the predictions of the networks reached at the end of each learning-rate cycle \cite{huang2017snapshot}. Model annealing keeps such an ensemble as its memory, uses it only as a teacher, and distills it into a single network within the same budget.

The adversarial component draws on two ideas. Generative adversarial networks cast learning as a minimax game between a generator and a discriminator \cite{goodfellow2014gan}, and adversarial belief matching transfers this game to distillation without data by training a generator to produce images on which student and teacher disagree \cite{micaelli2019zeroshot}. Virtual adversarial training searches a small neighborhood of every input for the perturbation that changes the model's own prediction the most and penalizes that change \cite{miyato2019vat}. Model annealing keeps the search local, as virtual adversarial training does, while the disagreement it maximizes is the one between the student and its consolidated past self.

Label smoothing also replaces a one-hot target by a distribution and improves both generalization and calibration \cite{muller2019ls}. The mass it spreads over the wrong classes is uniform, so it encodes nothing about which classes resemble each other. Comparing against it separates the value of structured, self-generated targets from the effect of target entropy as such.

\section{Model Annealing}

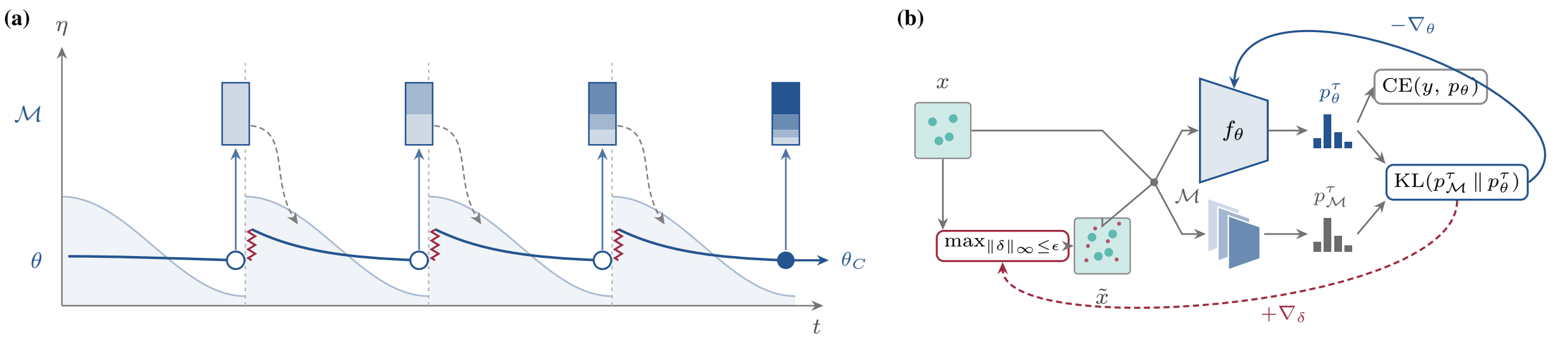
\begin{figure*}[t]
\centering
\resizebox{\textwidth}{!}{%
\begin{tikzpicture}[
  font=\footnotesize,
  >={Stealth[length=1.5mm,width=1.3mm]},
  snode/.style={circle, fill=white, draw=mnavy, line width=0.7pt, inner sep=0pt, minimum size=2.2mm},
  net/.style={trapezium, trapezium angle=72, shape border rotate=270, draw=#1, fill=#1!14, line width=0.7pt, minimum height=8.5mm, minimum width=5mm, inner sep=1pt},
  img/.style={rectangle, rounded corners=1.2pt, draw=black!45, line width=0.5pt, minimum size=7mm, inner sep=0pt},
  op/.style={rectangle, rounded corners=2.5pt, draw=#1, fill=white, line width=0.7pt, inner sep=2.6pt, font=\scriptsize},
  flow/.style={->, black!55, line width=0.6pt},
]
\def\W{9.2}
\colorlet{s1}{mnavy!22}
\colorlet{s2}{mnavy!42}
\colorlet{s3}{mnavy!68}
\colorlet{s4}{mnavy}
\foreach \k in {0,1,2,3} {
  \pgfmathsetmacro{\ea}{\k*\W/4}
  \pgfmathsetmacro{\eb}{(\k+1)*\W/4}
  \fill[mnavy!7] (\ea,0) -- plot[domain=\ea:\eb, samples=40, smooth] (\x, {0.12+1.25*0.5*(1+cos((\x-\ea)/(\W/4)*180))}) -- (\eb,0) -- cycle;
  \draw[mnavy!40, line width=0.6pt] plot[domain=\ea:\eb, samples=40, smooth] (\x, {0.12+1.25*0.5*(1+cos((\x-\ea)/(\W/4)*180))});
}
\draw[->, black!55, line width=0.5pt] (0,0) -- (\W+0.35,0) node[below left=0.5mm and -1mm, black!70] {$t$};
\draw[->, black!55, line width=0.5pt] (0,0) -- (0,3.25) node[above=0.2mm, black!70] {$\eta$};
\foreach \k in {1,2,3} {\draw[black!30, dash pattern=on 1.2pt off 1.4pt, line width=0.5pt] ({\k*\W/4},0) -- ({\k*\W/4},3.05);}
\foreach \k/\h in {0/0, 1/0.34, 2/0.34, 3/0.34} {
  \pgfmathsetmacro{\xs}{\k*\W/4+0.08}
  \pgfmathsetmacro{\xe}{(\k+1)*\W/4-0.12}
  \draw[mnavy, line width=0.9pt] (\xs,{0.62+\h}) .. controls ({\xs+0.55},{0.62+\h*0.2}) and ({\xe-0.9},0.6) .. (\xe,0.57);
}
\foreach \k in {1,2,3} {
  \draw[mburg, line width=0.7pt, decorate, decoration={zigzag, segment length=1.1mm, amplitude=0.45mm}] ({\k*\W/4+0.08},0.57) -- ({\k*\W/4+0.08},0.96);
}
\foreach \k in {1,2,3} {\node[snode] (t\k) at ({\k*\W/4-0.12},0.57) {};}
\node[circle, fill=mnavy, inner sep=0pt, minimum size=2.2mm] (t4) at ({\W-0.12},0.57) {};
\draw[->, mnavy, line width=0.8pt] ({\W-0.02},0.57) -- ({\W+0.42},0.57) node[right=0.2mm, font=\scriptsize] {$\theta_C$};
\def\bw{0.17}
\def\bh{0.78}
\def\by{2.02}
\foreach \k/\j/\lo/\hi in {1/1/0/1, 2/1/0/0.5, 2/2/0.5/1, 3/1/0/0.25, 3/2/0.25/0.5, 3/3/0.5/1, 4/1/0/0.125, 4/2/0.125/0.25, 4/3/0.25/0.5, 4/4/0.5/1} {
  \pgfmathsetmacro{\bx}{\k*\W/4-0.12}
  \fill[s\j] ({\bx-\bw},{\by+\lo*\bh}) rectangle ({\bx+\bw},{\by+\hi*\bh});
  \draw[white, line width=0.45pt] ({\bx-\bw},{\by+\hi*\bh}) -- ({\bx+\bw},{\by+\hi*\bh});
}
\foreach \k in {1,2,3,4} {
  \pgfmathsetmacro{\bx}{\k*\W/4-0.12}
  \draw[mnavy, line width=0.55pt] ({\bx-\bw},\by) rectangle ({\bx+\bw},{\by+\bh});
  \coordinate (m\k) at (\bx,\by);
}
\foreach \k in {1,2,3,4} {\draw[->, mnavy!80, line width=0.6pt] ({\k*\W/4-0.12},0.7) -- ([yshift=-0.4mm]m\k);}
\foreach \k in {1,2,3} {\draw[->, black!50, dash pattern=on 2pt off 1.2pt, line width=0.55pt] ([xshift=1.9mm,yshift=2.4mm]m\k) .. controls +(0.5,-0.05) and +(-0.35,0.55) .. ({\k*\W/4+0.66},1.02);}
\node[mnavy, anchor=east] at (-0.12,0.57) {$\theta$};
\node[mnavy, anchor=east] at (-0.12,{\by+0.5*\bh}) {$\mathcal{M}$};
\node[font=\bfseries\footnotesize, anchor=south west] at (-0.85,3.35) {(a)};

\begin{scope}[xshift=10.75cm]
\node[font=\bfseries\footnotesize, anchor=south west] at (-0.4,3.35) {(b)};
\node[img, fill=mjade!22] (x) at (0.3,2.2) {};
\foreach \i/\j in {0.17/2.31, 0.38/2.12, 0.25/2.07, 0.43/2.35} {\fill[mjade!75] (\i,\j) circle (0.55mm);}
\node[above=0.5mm of x, black!75] {$x$};
\node[op=mburg] (adv) at (1.05,0.75) {$\max_{\|\delta\|_\infty\le\epsilon}$};
\node[img, fill=mjade!22] (xt) at (2.3,0.75) {};
\foreach \i/\j in {2.17/0.86, 2.38/0.67, 2.25/0.62, 2.43/0.9} {\fill[mjade!75] (\i,\j) circle (0.55mm);}
\foreach \i/\j in {2.1/0.57, 2.48/0.58, 2.21/0.97, 2.37/0.81, 2.12/0.75, 2.51/1.03} {\fill[mburg!80] (\i,\j) circle (0.25mm);}
\node[below=0.5mm of xt, black!75] {$\tilde{x}$};
\draw[flow] (x.south) -- (x.south |- adv.north);
\draw[flow] (adv.east) -- (xt.west);
\coordinate (j) at (2.95,1.55);
\fill[black!55] (j) circle (0.5mm);
\draw[black!55, line width=0.6pt] (x.east) -- (2.3,2.2) -- (j);
\draw[black!55, line width=0.6pt] (xt.north) -- (2.3,1.0) -- (j);
\node[net=mnavy] (fs) at (3.95,2.2) {$f_\theta$};
\foreach \dx/\dy/\c in {-0.13/0.11/s1, 0/0/s2, 0.13/-0.11/s3} {
  \filldraw[fill=\c, draw=white, line width=0.5pt] ({3.95+\dx-0.2},{0.9+\dy-0.34}) -- ({3.95+\dx+0.2},{0.9+\dy-0.2}) -- ({3.95+\dx+0.2},{0.9+\dy+0.2}) -- ({3.95+\dx-0.2},{0.9+\dy+0.34}) -- cycle;
}
\node[black!65, font=\scriptsize, anchor=south east] at (3.66,1.17) {$\mathcal{M}$};
\draw[flow] (j) -- (3.3,2.2) -- (fs.west);
\draw[flow] (j) -- (3.3,0.9) -- (3.6,0.9);
\foreach \c/\yy in {mnavy/2.2, mteach/0.9} {
  \foreach \b/\hh in {0/0.12, 1/0.42, 2/0.2, 3/0.08} {\fill[\c] ({4.95+\b*0.13},{\yy-0.22}) rectangle ({4.95+\b*0.13+0.09},{\yy-0.22+\hh});}
}
\draw[flow] (fs.east) -- (4.88,2.2);
\draw[flow] (4.33,0.9) -- (4.88,0.9);
\node[mnavy, font=\scriptsize, anchor=south] at (5.17,2.42) {$p_\theta^{\tau}$};
\node[black!65, font=\scriptsize, anchor=south] at (5.17,1.12) {$p_{\mathcal{M}}^{\tau}$};
\node[op=mnavy] (kl) at (6.75,1.55) {$\mathrm{KL}(p_{\mathcal{M}}^{\tau}\,\|\,p_\theta^{\tau})$};
\node[op=black!45] (ce) at (6.42,2.75) {$\mathrm{CE}(y,\,p_\theta)$};
\draw[flow] (5.5,2.2) -- (kl.north west);
\draw[flow] (5.5,0.9) -- (kl.south west);
\draw[flow] (5.5,2.3) -- (ce.west);
\draw[->, mburg, line width=0.75pt, dash pattern=on 2.2pt off 1.3pt] (kl.south) .. controls +(0,-1.35) and +(0,-1.0) .. node[pos=0.42, below=0.2mm, font=\scriptsize] {$+\nabla_{\delta}$} (adv.south);
\draw[->, mnavy, line width=0.75pt] (kl.east) .. controls +(1.25,0.9) and +(0.1,1.75) .. (fs.north);
\node[mnavy, font=\scriptsize, anchor=south] at (6.2,3.3) {$-\nabla_{\theta}$};
\end{scope}
\end{tikzpicture}%
}
\caption{Model annealing. (a) Reheated episodes and the memory of cooled snapshots. (b) One relearning step.}
\label{fig:concept}
\end{figure*}

Model annealing trains a student network $f_\theta$ and keeps a memory $\mathcal{M}$ of the states it has cooled into. The budget of $T$ optimizer steps is divided into $C$ episodes of equal length. In every episode the learning rate is warmed up to the same peak $\eta_{\max}$ and then decays along a cosine to nearly zero (Fig.~\ref{fig:concept}(a)). Because the learning rate sets the scale of the gradient noise, raising it again at the start of an episode reheats the system, and the weights become free to leave the basin in which the previous episode came to rest. Only the cooled end of an episode enters the memory, a step we call consolidation. A network sampled at a high learning rate makes a poor teacher: for a ResNet-18 on CIFAR-10, the mean gradients of consecutive epochs at a high learning rate were found to be almost orthogonal \cite{ruzbahani2026gradreason}, so the weights there move mostly with mini-batch noise. The first episode is ordinary supervised training. Each later episode relearns from the labels, from the memory and from adversarial dreams, and every episode, the first included, ends by adding its final state to the memory.

\subsection{Relearning from a Memory of Past Selves}

After $c$ episodes the memory holds the snapshots $\theta_1,\dots,\theta_c$ and predicts
\begin{equation}
p^{\tau}_{\mathcal{M}}(x)=\sum_{j=1}^{c} w_{c,j}\,p^{\tau}_{\theta_j}(x),\quad w_{c,j}=\begin{cases}(1-\mu)^{c-1}, & j=1,\\ \mu\,(1-\mu)^{c-j}, & j\geq2,\end{cases}
\label{eq:memory}
\end{equation}
where $p^{\tau}$ is the softmax at temperature $\tau$. The newest snapshot carries weight $\mu$, and the influence of older ones decays geometrically without ever vanishing. During the next episode the student minimizes
\begin{equation}
\begin{aligned}
\mathcal{L}(\theta)={}&(1-\alpha)\,\mathrm{CE}\big(y,p_\theta(x)\big)+\alpha\,\tau^2\,\mathrm{KL}\big(p^{\tau}_{\mathcal{M}}(x)\,\|\,p^{\tau}_\theta(x)\big)\\
&+\beta\,\tau^2\,\mathrm{KL}\big(p^{\tau}_{\mathcal{M}}(\tilde{x})\,\|\,p^{\tau}_\theta(\tilde{x})\big),
\end{aligned}
\label{eq:loss}
\end{equation}
where $x$ is an augmented training image with label $y$ and $\tilde{x}$ is a dream. The memory sees the same augmented view as the student, so it always speaks about the image the student is looking at. The supervision for a training image is therefore the mixture $q=(1-\alpha)\,e_y+\alpha\,p^{\tau}_{\mathcal{M}}(x)$ instead of the one-hot vector $e_y$. Its entropy is zero for cross-entropy and constant for label smoothing; here it varies from image to image and is shaped by what the network learned in all earlier episodes.

The weights in (\ref{eq:memory}) are the ones a Reptile step $\phi_c=\phi_{c-1}+\mu\,(\theta_c-\phi_{c-1})$ \cite{nichol2018reptile} would assign to the snapshots if they were averaged in weight space. We average predictions instead. A weight average is meaningful when the averaged points share one basin, as successive iterates of a single trajectory at a small learning rate do \cite{izmailov2018swa}. Reheating, however, carries the weights far enough that the straight path between consecutive snapshots can cross a ridge of high loss \cite{frankle2020linear}. Predictions can be averaged regardless, as snapshot ensembles do with the networks found at the end of their learning-rate cycles \cite{huang2017snapshot}. With $\mu=1$ the memory reduces to the latest snapshot, the teacher of snapshot distillation \cite{yang2019snapshot}. The memory serves training only. The network returned at the end is the last student, a single model whose inference cost equals that of every baseline.

\subsection{Dreams}

Dreams are produced by an adversary that searches an $\ell_\infty$ ball around a training image for the point where the student departs most from its memory,
\begin{equation}
\tilde{x}=\Pi_{\epsilon}\big[x+\delta_0+\tfrac{5}{4}\epsilon\,\mathrm{sign}(g)\big],
\label{eq:dream}
\end{equation}
where $g=\nabla_{\delta}\,\mathrm{KL}\big(p^{\tau}_{\mathcal{M}}(x+\delta_0)\,\|\,p^{\tau}_\theta(x+\delta)\big)$ is evaluated at $\delta=\delta_0$, $\delta_0$ is drawn uniformly from the ball, the memory's prediction is held fixed during the ascent step, and $\Pi_\epsilon$ projects back onto the ball and the valid pixel range. The adversary ascends the discrepancy that the third term of (\ref{eq:loss}) descends. The two players form a game in the spirit of generative adversarial training \cite{goodfellow2014gan}, played locally around real images rather than in the output space of a generator. The memory labels every dream with its tempered distribution, so the student receives supervision at points that lie outside the training set and were selected to maximize the disagreement between its current knowledge and what it knew before. Algorithm~\ref{alg:ma} summarizes the procedure.

\begin{algorithm}[t]
\caption{Model annealing}
\label{alg:ma}
\begin{algorithmic}[1]
\Require data $\mathcal{D}$, steps $T$, episodes $C$, temperature $\tau$, weights $\alpha,\beta$, radius $\epsilon$, memory rate $\mu$
\State initialize $\theta$ and an empty memory $\mathcal{M}$
\For{$c=1,\dots,C$}
  \For{$T/C$ steps, learning rate warmed up to $\eta_{\max}$ and cosine-decayed}
    \State draw and augment a mini-batch $(x,y)$ from $\mathcal{D}$
    \If{$c=1$}
      \State $\ell\gets\mathrm{CE}(y,p_\theta(x))$
    \Else
      \State form dreams $\tilde{x}$ for half of the batch by (\ref{eq:dream})
      \State $\ell\gets$ objective (\ref{eq:loss})
    \EndIf
    \State take an optimizer step on $\ell$
  \EndFor
  \State add a frozen copy of $\theta$ to $\mathcal{M}$ and reweight it by (\ref{eq:memory})
\EndFor
\State \Return $\theta$
\end{algorithmic}
\end{algorithm}

\section{Experimental Setup}\label{sec:setup}

The evaluation spans three image classification benchmarks of increasing resolution, four architecture families and four amounts of labelled data, with every comparison paired on identical data, initialization and batch order.

CIFAR-100 \cite{krizhevsky2009cifar} contains 50{,}000 training and 10{,}000 test images of 100 classes at $32\times32$ pixels. Tiny-ImageNet is a 200-class subset of ImageNet \cite{russakovsky2015imagenet} with 100{,}000 training images at $64\times64$ pixels; its test labels are not public, so its official validation split serves as the test set. Imagewoof\footnote{\url{https://github.com/fastai/imagenette}} draws ten visually similar dog breeds from ImageNet, with 9{,}025 training and 3{,}929 test images that we resize to 144 pixels and crop to 128. From each official training set a fixed, stratified 10\% is held out for validation, and nested class-balanced subsets of 5\%, 10\%, 25\% and 100\% of the remainder form the training sets, the full set being omitted for Tiny-ImageNet. Test images are evaluated once, after training. Every decision that depends on data, from learning rates to checkpoint selection and temperature scaling, is made on the validation split, and normalization statistics come from each run's own training subset. A search for exact and near-duplicate images finds 33, 7 and 3 test images of CIFAR-100, Tiny-ImageNet and Imagewoof with a counterpart in the training data; on the de-duplicated test sets no paired difference in top-1 accuracy between model annealing and a baseline changes sign. Out-of-distribution detection uses SVHN \cite{netzer2011svhn} as far-distribution data for all three benchmarks and the disjoint Imagenette classes as near-distribution data for Imagewoof.

All networks are trained from scratch. ResNet-18 \cite{he2016resnet} stands for convolutions with batch normalization, ConvNeXt \cite{liu2022convnext} in an 8.6\,M-parameter configuration for depthwise convolutions with layer normalization, ViT-Tiny \cite{dosovitskiy2021vit} for self-attention, and an eight-block MLP-Mixer \cite{tolstikhin2021mixer} of width 384 for pure token and channel mixing. Their stems are adapted so that ResNet-18 sees a $32\times32$ feature map, ConvNeXt a $16\times16$ map and the two token-based models 64 patches, whatever the input resolution. ResNet-18 is optimized with Nesterov SGD (momentum 0.9, weight decay $5\times10^{-4}$) and the other families with AdamW \cite{loshchilov2019adamw} (weight decay 0.05, gradient norm clipped at 1, drop-path rate 0.1). Batches hold 256 images, the learning rate warms up linearly over 5\% of training and then follows a cosine decay, and a run lasts the larger of 100 epochs and 4{,}000 steps so that small subsets still receive enough updates. The base learning rate of each family was selected by a cross-entropy sweep with one cosine cycle on the CIFAR-100 validation split at 10\% and 25\% of the data and is shared by every method: 1.6 for ResNet-18, the largest value tried, and $2.5\times10^{-4}$, $5\times10^{-4}$ and $10^{-3}$ for ConvNeXt, ViT and MLP-Mixer, the value for ConvNeXt being the smallest tried. The sweep used a slightly longer budget than the main runs, 120 epochs and at least 5{,}000 steps. Augmentation consists of reflection-padded random crops and horizontal flips at 32 and 64 pixels, and of random resized crops with scale between 0.35 and 1 plus flips at 128 pixels.

Model annealing is compared with cross-entropy training, label smoothing with $\epsilon=0.1$ \cite{muller2019ls}, SGDR with four cycles \cite{loshchilov2017sgdr}, and snapshot distillation \cite{yang2019snapshot} using four cycles and the same temperature and distillation weight as model annealing. Ablations on CIFAR-100 with 25\% of the data, most of them repeated at 10\%, replace the memory by the latest snapshot, remove the dreams, average the snapshots in weight space instead of function space, add Gaussian weight noise at every reheat, halve the peak learning rate from one episode to the next, or make every episode twice or half as long as the one before it; the front-loaded schedule is also run with SGDR. A cross-entropy run with twice the number of steps serves as a compute reference, and a diagnostic variant distills from a cross-entropy teacher trained on the entire training pool, which separates the quality of the teacher from the act of distillation. Model annealing uses $C=4$, $\tau=4$, $\alpha=\beta=\tfrac12$, $\mu=\tfrac12$ and $\epsilon=4/255$, with dreams on half of every batch. The adversary evaluates the student in inference mode. These values were fixed before any experiment; the form of the memory and the use of dreams without weight noise were chosen on the validation split of CIFAR-100 at 10\% and 25\% data. Beyond top-1 accuracy we report negative log-likelihood, expected calibration error with 15 bins \cite{guo2017calibration}, the area under the ROC curve for out-of-distribution detection with the maximum softmax probability \cite{hendrycks2017baseline}, and accuracy under five synthetic corruptions at three severities in the spirit of \cite{hendrycks2019corruptions}. Paired differences are tested across configurations with the Wilcoxon signed-rank test, with Holm's correction over all comparisons of a table.

Training on the GPU is not bitwise deterministic, so identical computations can end in different networks. The first episode is the same computation for model annealing and snapshot distillation, from the same initialization on the same batches, yet the validation accuracies the two reach at its end differ with a standard deviation of 0.7 points over the 44 configurations, of 0.3 points for the AdamW families, and by at most 3.4 points. We use this spread as a yardstick for which differences between methods are meaningful.

\section{Results}

\begin{figure*}[t]
\centering
\includegraphics[width=\textwidth]{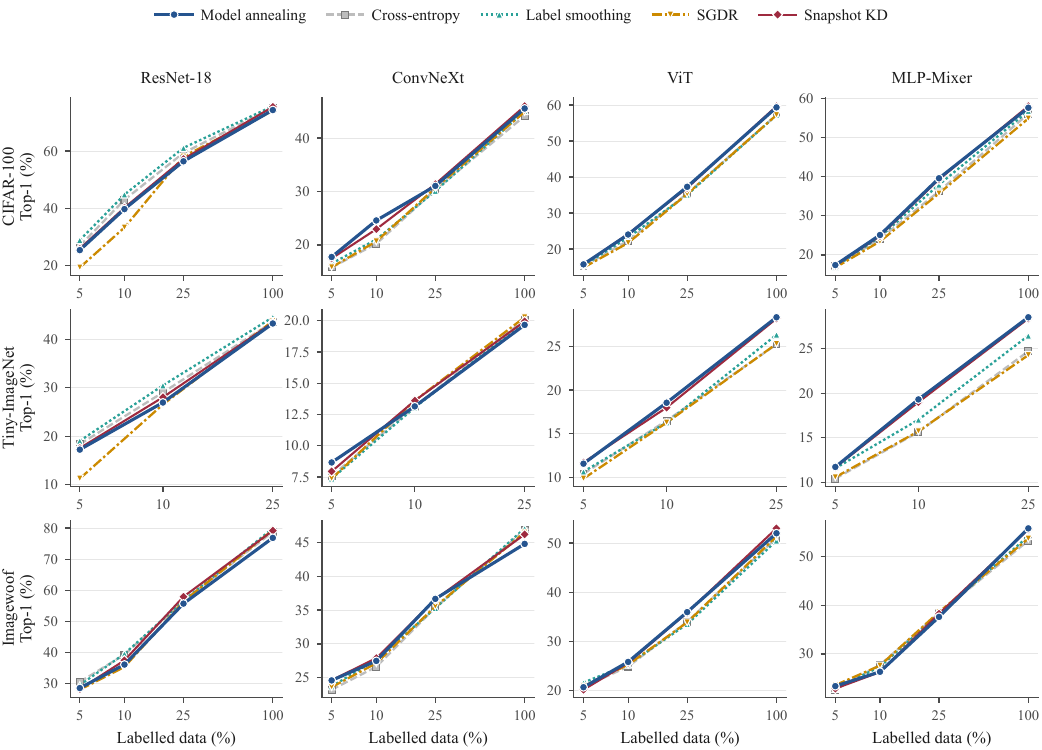}
\caption{Test top-1 accuracy against the amount of labelled data.}
\label{fig:accuracy}
\end{figure*}

\begin{figure*}[t]
\centering
\includegraphics[width=\textwidth]{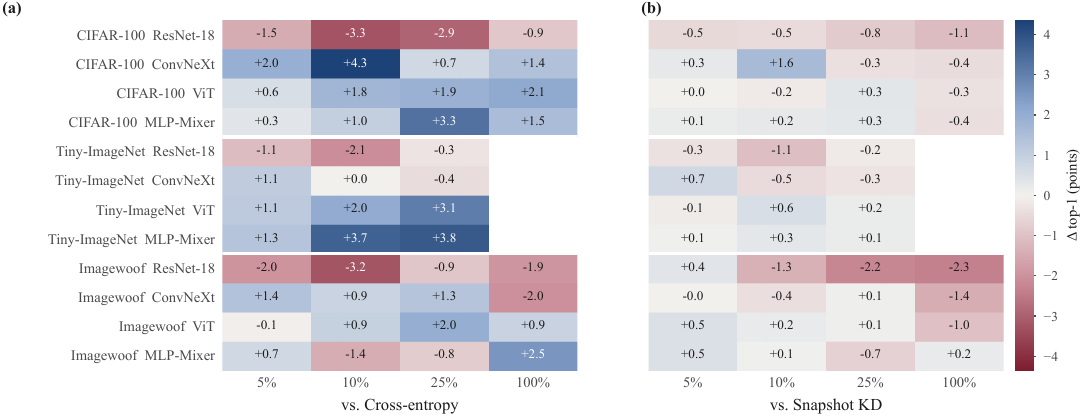}
\caption{Top-1 gain of model annealing over (a) cross-entropy and (b) snapshot distillation, per configuration.}
\label{fig:gain}
\end{figure*}

Table~\ref{tab:main} reports the test accuracy of every configuration, Fig.~\ref{fig:accuracy} plots it against the amount of labelled data, Fig.~\ref{fig:gain} maps the gain of model annealing configuration by configuration, and Table~\ref{tab:summary} condenses the paired differences between model annealing and each baseline. The three families trained with AdamW and ResNet-18, the only family trained with SGD, respond to model annealing in opposite directions, so we report them separately.

\begin{table*}[t]
\centering
\caption{Test top-1 accuracy (\%). Bold: best method per model and column.}
\label{tab:main}
\setlength{\tabcolsep}{3pt}
\resizebox{\textwidth}{!}{%
\begin{tabular}{llccccccccccc}
\toprule
 & & \multicolumn{4}{c}{CIFAR-100} & \multicolumn{3}{c}{Tiny-ImageNet} & \multicolumn{4}{c}{Imagewoof}\\
\cmidrule(lr){3-6}\cmidrule(lr){7-9}\cmidrule(lr){10-13}
Model & Method & 5\% & 10\% & 25\% & 100\% & 5\% & 10\% & 25\% & 5\% & 10\% & 25\% & 100\%\\
\midrule
\multirow{5}{*}{ResNet-18} & Model annealing & 25.4 & 39.7 & 56.4 & 74.3 & 17.2 & 27.0 & 43.3 & 28.5 & 36.1 & 55.7 & 76.9\\
 & Cross-entropy & 26.8 & 43.0 & 59.3 & 75.2 & 18.3 & 29.0 & 43.6 & \textbf{30.5} & 39.3 & 56.6 & 78.8\\
 & Label smoothing & \textbf{28.9} & \textbf{44.8} & \textbf{61.1} & \textbf{75.7} & \textbf{19.0} & \textbf{30.5} & \textbf{44.6} & 29.5 & \textbf{39.5} & 56.6 & \textbf{79.9}\\
 & SGDR & 19.3 & 33.3 & 58.1 & 74.1 & 11.2 & 26.6 & 43.8 & 28.0 & 35.3 & 56.4 & 79.4\\
 & Snapshot KD & 25.9 & 40.2 & 57.2 & 75.4 & 17.5 & 28.0 & 43.5 & 28.1 & 37.4 & \textbf{57.9} & 79.2\\
\midrule
\multirow{5}{*}{ConvNeXt} & Model annealing & \textbf{17.7} & \textbf{24.6} & 31.0 & 45.5 & \textbf{8.7} & 13.2 & 19.7 & 24.6 & 27.5 & \textbf{36.7} & 44.8\\
 & Cross-entropy & 15.8 & 20.2 & 30.3 & 44.1 & 7.6 & 13.1 & 20.1 & 23.2 & 26.6 & 35.4 & 46.8\\
 & Label smoothing & 16.5 & 21.0 & 30.1 & 45.1 & 7.4 & 13.1 & 19.9 & 23.7 & 27.7 & 35.3 & \textbf{47.1}\\
 & SGDR & 15.8 & 20.7 & 30.7 & 44.9 & 7.4 & \textbf{13.6} & \textbf{20.3} & 23.6 & 27.2 & 35.5 & 46.4\\
 & Snapshot KD & 17.5 & 22.9 & \textbf{31.4} & \textbf{46.0} & 8.0 & \textbf{13.6} & 19.9 & \textbf{24.6} & \textbf{27.9} & 36.6 & 46.2\\
\midrule
\multirow{5}{*}{ViT} & Model annealing & \textbf{15.7} & 24.0 & \textbf{37.3} & 59.4 & 11.6 & \textbf{18.6} & \textbf{28.3} & 20.7 & \textbf{25.8} & \textbf{36.0} & 52.0\\
 & Cross-entropy & 15.2 & 22.3 & 35.4 & 57.2 & 10.5 & 16.5 & 25.2 & 20.7 & 24.9 & 34.0 & 51.1\\
 & Label smoothing & 15.4 & 23.2 & 35.1 & 57.2 & 10.7 & 16.3 & 26.3 & \textbf{21.6} & 25.3 & 33.5 & 50.5\\
 & SGDR & 14.8 & 21.7 & 35.3 & 57.2 & 9.9 & 16.3 & 25.3 & 20.9 & 25.3 & 33.9 & 51.4\\
 & Snapshot KD & 15.7 & \textbf{24.2} & 37.0 & \textbf{59.7} & \textbf{11.7} & 18.0 & 28.1 & 20.2 & 25.6 & 35.9 & \textbf{53.0}\\
\midrule
\multirow{5}{*}{MLP-Mixer} & Model annealing & \textbf{17.4} & \textbf{25.1} & \textbf{39.6} & 57.7 & \textbf{11.7} & \textbf{19.3} & \textbf{28.5} & 23.4 & 26.3 & 37.6 & \textbf{55.8}\\
 & Cross-entropy & 17.1 & 24.1 & 36.3 & 56.1 & 10.4 & 15.6 & 24.7 & 22.6 & \textbf{27.7} & 38.4 & 53.3\\
 & Label smoothing & 17.0 & 24.6 & 38.0 & 56.8 & 11.6 & 17.0 & 26.5 & 22.8 & 27.4 & 38.2 & 54.2\\
 & SGDR & 16.8 & 23.4 & 35.7 & 54.9 & 10.6 & 15.7 & 24.3 & \textbf{23.6} & 27.6 & \textbf{38.6} & 53.7\\
 & Snapshot KD & 17.3 & 24.9 & 39.3 & \textbf{58.1} & 11.7 & 19.0 & 28.4 & 22.9 & 26.3 & 38.3 & 55.6\\
\bottomrule
\end{tabular}%
}
\end{table*}

\begin{table}[t]
\centering
\caption{Model annealing minus each baseline, averaged over configurations (NLL in nats, others in points). $^{*}$: $p<0.05$, Wilcoxon test with Holm correction. Wins: configurations with higher top-1 accuracy.}
\label{tab:summary}
\setlength{\tabcolsep}{2.4pt}
\resizebox{\columnwidth}{!}{%
\begin{tabular}{llccccccc}
\toprule
Families & vs. & Top-1 & NLL & ECE & Corr. & PGD & OOD & Wins\\
\midrule
\multirow{4}{*}{\shortstack[l]{ConvNeXt,\\ViT,\\Mixer}} & Cross-entropy & +1.31$^{*}$ & $-$1.58$^{*}$ & $-$18.7$^{*}$ & +1.27$^{*}$ & +4.0$^{*}$ & +0.5 & 28/33\\
 & Label smoothing & +0.90$^{*}$ & +0.15 & +3.6 & +1.00$^{*}$ & +3.0$^{*}$ & +0.1 & 27/33\\
 & SGDR & +1.30$^{*}$ & $-$1.42$^{*}$ & $-$17.1$^{*}$ & +1.18$^{*}$ & +3.9$^{*}$ & +0.5 & 26/33\\
 & Snapshot KD & +0.01 & $-$0.26$^{*}$ & $-$5.5$^{*}$ & +0.22$^{*}$ & +1.8$^{*}$ & +0.1 & 20/33\\
\midrule
\multirow{4}{*}{\shortstack[l]{ResNet-18}} & Cross-entropy & $-$1.82$^{*}$ & +0.37 & +2.6 & +0.35 & +9.0$^{*}$ & +2.3 & 0/11\\
 & Label smoothing & $-$2.69$^{*}$ & +0.68 & +15.6 & $-$0.36 & +7.1 & +2.4 & 0/11\\
 & SGDR & +1.37 & $-$0.38 & $-$3.0 & +1.31 & +10.6$^{*}$ & +3.1 & 7/11\\
 & Snapshot KD & $-$0.90 & $-$0.05 & $-$0.9 & +0.06 & +5.5$^{*}$ & $-$0.3 & 1/11\\
\bottomrule
\end{tabular}%
}
\end{table}

\subsection{Accuracy, Calibration and Robustness}

For ConvNeXt, ViT and MLP-Mixer, model annealing is more accurate than cross-entropy training in 28 of 33 configurations, by 1.3 points on average and by up to 4.3 points for ConvNeXt on CIFAR-100 with 10\% of the data. It also exceeds label smoothing by 0.9 points and SGDR, whose learning-rate schedule it shares, by 1.3 points. Snapshot distillation is the closest competitor at 1.3 points above cross-entropy, and in accuracy model annealing is on par with it, with an average difference close to zero (Table~\ref{tab:summary}). Model annealing is ahead of it in 19 of 27 configurations with at most a quarter of the data (+0.14 points on average, less than the spread between identical computations described in Section~\ref{sec:setup}) and behind it when all the data are used ($-$0.56 points). Model annealing is better calibrated and more robust than snapshot distillation, with a negative log-likelihood lower by 0.26 nats, an expected calibration error lower by 5.5 points, and accuracy higher by 0.2 points under corruptions and by 1.8 points under the PGD attack described below, all four differences significant (Table~\ref{tab:summary}).

Relearning from soft targets changes more than the top prediction. Top-5 accuracy rises by 3.3 points over cross-entropy and by 5.0 points over label smoothing. The negative log-likelihood falls by 1.58 nats relative to cross-entropy and remains 0.16 nats lower once both models are temperature-scaled on the validation split, so the improvement is more than a rescaling of confidence. The expected calibration error drops by 18.7 points, although label smoothing is better calibrated before temperature scaling in 20 of 33 configurations. The two kinds of soft target also differ in what they encode. On CIFAR-100 a uniform spread would place 4\% of the non-target probability on the four other classes of the true superclass. Cross-entropy places 23.3\% there, label smoothing 11.9\%, and model annealing 23.4\%: label smoothing weakens the similarity structure among the wrong classes, and a memory of past selves preserves it.

Robustness improves for every family. Accuracy under corruptions exceeds that of cross-entropy in 36 of 44 configurations, by 1.0 points on average, and accuracy under a ten-step PGD attack with $\epsilon=1/255$ in 42 of 44, by 5.2 points. For ResNet-18 the gain under attack holds (+9.0 points) even though its clean accuracy falls, while the gain under corruptions shrinks to +0.3 points and is not significant. The AUROC for detecting SVHN images as out-of-distribution changes by +0.9 points, which is not significant.

ResNet-18 is the exception in clean accuracy. Model annealing trails cross-entropy by 1.8 points and label smoothing, the most accurate method for this family in 9 of 11 configurations, by 2.7 points. Snapshot distillation also falls behind cross-entropy ($-$0.9 points). The analysis below traces both losses to the reheated schedule rather than to relearning.

\subsection{Where Relearning Helps, and Why}

\begin{figure*}[t]
\centering
\includegraphics[width=\textwidth]{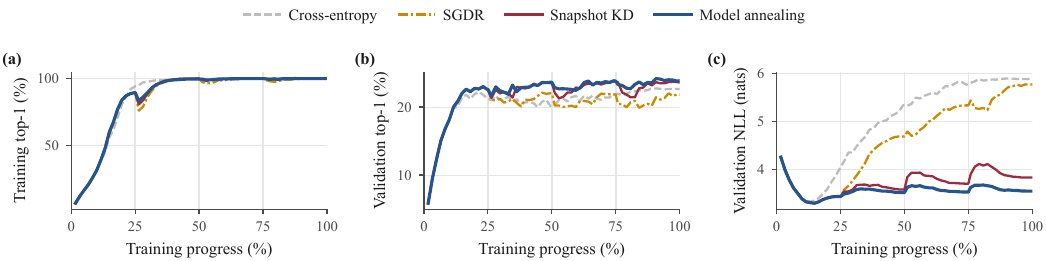}
\caption{Training of ViT on CIFAR-100 with 10\% of the labels. Vertical lines mark the reheats.}
\label{fig:dynamics}
\end{figure*}

\begin{figure*}[t]
\centering
\includegraphics[width=\textwidth]{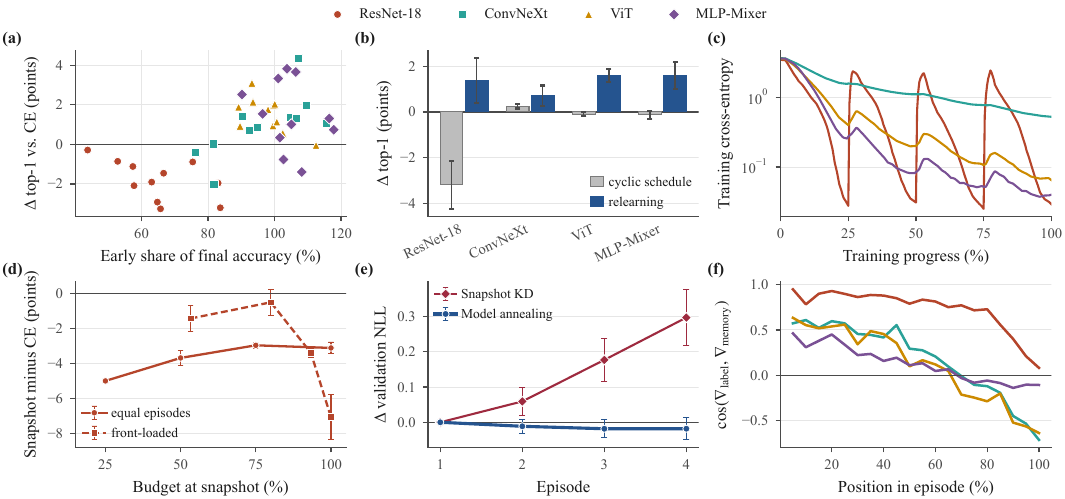}
\caption{(a) Gain over cross-entropy against its early share of final accuracy. (b) Schedule and relearning effects. (c) Training cross-entropy under model annealing. (d) ResNet-18 snapshots relative to cross-entropy. (e) Validation NLL relative to the first episode. (f) Cosine between label and memory gradients. Error bars: standard errors over configurations.}
\label{fig:where}
\end{figure*}

Fig.~\ref{fig:dynamics} follows a single configuration through training. Cross-entropy fits its training images within the first half of the budget, its validation accuracy stops rising, and its validation NLL climbs for the rest of the run, while that of snapshot distillation steps up at every reheat and that of model annealing stays close to its first-episode level. The first episode of model annealing ends after a quarter of the budget, and its snapshot becomes the first teacher. Fig.~\ref{fig:where}(a) relates the gain over cross-entropy to the share of its final validation accuracy that a cross-entropy run has reached at that point. For the AdamW families the share lies between 76\% and 118\%, and in 18 of 33 configurations the accuracy after one quarter is higher than at the end. In these configurations the remaining three quarters of the budget buy memorization rather than accuracy, and relearning puts them to better use. ResNet-18 has reached 44\% to 84\% by then, so the two ranges overlap between 76\% and 84\%. Across all configurations, share and gain have a Spearman correlation of 0.54 ($p<0.001$), but across the configurations of the AdamW families only +0.07. The overall correlation thus rests on the difference between ResNet-18 and the AdamW families, and among the AdamW configurations the share does not predict the gain.

Model annealing differs from cross-entropy training in two respects, the reheated schedule and relearning from the memory. SGDR has the first without the second, so SGDR minus cross-entropy isolates the schedule and model annealing minus SGDR isolates relearning (Fig.~\ref{fig:where}(b)). Relearning raises accuracy for every family, by 0.7 to 1.6 points on average and in 33 of 44 configurations. The schedule is close to neutral for the AdamW families ($-$0.1 to +0.3 points) but costs ResNet-18 3.2 points on average and up to 9.7 points with 10\% of the data. The difference lies in what a reheat does to each network (Fig.~\ref{fig:where}(c)). When the SGD learning rate of ResNet-18 returns to its peak, the training cross-entropy of the network jumps from 0.13 to 2.54 nats, against 3.76 at the start of training, and its accuracy on training batches, 98\% on average before the reheat, falls to between 26\% and 45\%. The AdamW families, whose accuracy on training batches averages 78\% before a reheat, lose at most 7 points of it per reheat, averaged over the reheats of each configuration. Between consecutive snapshots the weights of ResNet-18 move by 42\% to 140\% of their norm, against 6\% to 32\% for the other families. Every episode of this network thus begins far below the fit the previous one reached, and its snapshot records a short training run rather than a refinement of the previous one.

Front-loading the schedule tests this account (Fig.~\ref{fig:where}(d)). When the four episodes take 8/15, 4/15, 2/15 and 1/15 of the budget, the first snapshot of ResNet-18 comes within 1.4 points of the validation accuracy of a full cross-entropy run, against 5.0 points for the first of four equal episodes. The second episode, still long enough to recover from its reheat, adds 0.9 points in 2 of 2 configurations. The two short episodes that follow end before the network has recovered, and the last snapshot falls 6.5 points below the second. Relearning thus helps ResNet-18 too, but only within episodes that outlast the recovery from a reheat. The AdamW families hardly register a reheat, and for them front-loading merely postpones relearning ($-$0.88 points against equal episodes).

Snapshot distillation shares the schedule of model annealing but relearns from the latest snapshot alone, a snapshot that was itself trained to imitate its predecessor. With at most a quarter of the data this chain compounds overconfidence (Fig.~\ref{fig:where}(e)). For the AdamW families the validation NLL of its student rises by 0.06, 0.18 and 0.30 nats over episodes two to four, and the accuracy gained since the first episode stalls at 1.6 points after episodes three and four. Under model annealing the NLL never moves by more than 0.02 nats on average from its first-episode value, while the accuracy keeps rising to 1.7 points. The first snapshot never learned from a teacher and keeps a share of every later memory, which may anchor the targets against this drift. Snapshot distillation also lacks the dreams, and the ablations on CIFAR-100 at 10\% and 25\% of the data separate the memory from the dreams. From the first episode to the last, the validation NLL of the AdamW families changes by +0.18 nats under snapshot distillation, by +0.06 when the dreams alone are added to it, by +0.03 when the memory alone replaces its teacher, and by $-$0.04 under model annealing. On average, the memory and the dreams each curb most of the drift, and together they remove it. With all the data each snapshot clearly improves on its predecessor, the latest one appears to be the better teacher, and model annealing may pay for down-weighting it.

The gradient probe in Fig.~\ref{fig:where}(f) shows when the memory tells the student something the labels do not. On a fixed set of training images, the label and memory gradients of the AdamW families agree at the start of an episode (cosine 0.51) and oppose each other at its end ($-$0.36). Once the cooling student starts to fit the labels more sharply than its memory does, the memory holds it back. For ResNet-18 the two gradients remain aligned (0.89 early in the episode), and their cosine falls below half of that value only at about 90\% of the episode: the reheated student spends most of the episode recovering what it lost, and the memory repeats what the labels already say.

\subsection{Ablations and Cost}

\begin{table}[t]
\centering
\caption{Ablations on CIFAR-100: variant minus model annealing. Units and $^{*}$ as in Table~\ref{tab:summary}.}
\label{tab:ablation}
\setlength{\tabcolsep}{3pt}
\resizebox{\columnwidth}{!}{%
\begin{tabular}{lccccc}
\toprule
 & \multicolumn{2}{c}{Top-1} & & & \\
\cmidrule(lr){2-3}
Variant & AdamW & SGD & NLL & Corr. & PGD\\
\midrule
Latest snapshot as teacher & $-$0.18 & $-$0.24 & +0.082 & +0.06 & +0.4\\
Without dreams & $-$0.06 & +0.88 & +0.043 & $-$0.53 & $-$2.7\\
Weight-space memory & $-$0.28 & $-$3.17 & +0.033 & +0.47 & +0.6\\
With weight noise & $-$0.12 & +0.88 & $-$0.002 & +0.06 & $-$0.2\\
Halving reheats & $-$1.54 & +0.61 & +0.056 & $-$0.61 & $-$2.4\\
Growing episodes & $-$2.39 & $-$2.44 & +0.188 & +1.02 & $-$0.0\\
Front-loaded episodes & $-$0.88 & $-$4.60 & +0.777 & $-$0.35 & $-$0.4\\
\midrule
Snapshot KD & $-$0.10 & +0.71 & +0.176$^{*}$ & $-$0.43$^{*}$ & $-$2.2$^{*}$\\
Cross-entropy, $2\times$ steps vs.\ cross-entropy & $-$0.36 & +1.63 & +0.740 & $-$0.30 & $-$0.2\\
\bottomrule
\end{tabular}%
}
\end{table}

\begin{figure}[t]
\centering
\includegraphics[width=\columnwidth]{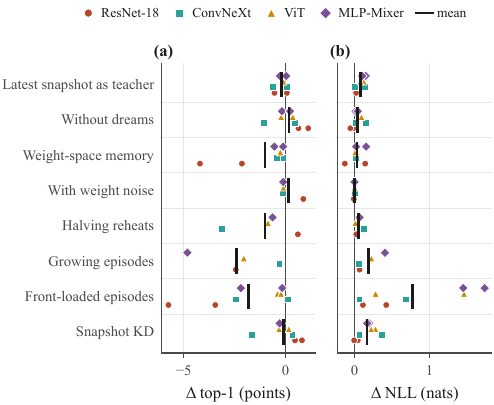}
\caption{Ablations on CIFAR-100, variant minus model annealing per configuration. Bars mark means.}
\label{fig:ablation}
\end{figure}

Table~\ref{tab:ablation} and Fig.~\ref{fig:ablation} isolate the components on CIFAR-100. For the AdamW families no variant beats model annealing on average, and the schedule is what matters most: halved peaks, growing episodes and front-loaded episodes cost them 0.88 to 2.39 points. Swapping the memory for the latest snapshot or dropping the dreams moves their top-1 accuracy by at most 0.18 points, yet both raise the NLL, and over all four families the loss of the dreams also costs 2.7 points under PGD and 0.53 under corruptions. Weight noise at the reheats leaves them essentially unchanged. A memory averaged in weight space is the weakest teacher. It falls 1.4 points short of the student it is built from, whereas the function-space memory stays within 0.1 points of its student and is better calibrated by 0.08 nats, and it costs the AdamW families 0.28 points and ResNet-18 3.17, since the reheats of this network carry its snapshots far apart. ResNet-18 also suffers most from short episodes, and in this network the dreams trade 0.88 points of clean accuracy for 4.9 points under PGD.

Model annealing takes 2.7 times as long to train as cross-entropy (median over configurations) and snapshot distillation 1.3 times. The dreams and the forward passes through up to three memory members account for the difference, and inference costs the same for every method. Doubling the steps of cross-entropy changes accuracy by $-$0.4 points for the AdamW families, with gains in 3 of 6 configurations, and by +1.6 for ResNet-18, the family that saturates latest, with gains in 2 of 2. For the AdamW families more plain training thus buys no accuracy, and over all four families it raises the NLL (Table~\ref{tab:ablation}). Distilling from a cross-entropy teacher trained on the entire training pool raises accuracy at 10\% of the CIFAR-100 data by 7.8 points over cross-entropy for the AdamW families and by 17.6 for ResNet-18, and by 5.4 and 20.9 points over model annealing. This teacher has learned from labels the student never sees, so it marks what a better teacher is worth rather than competing with model annealing, and the large gain of ResNet-18 shows that learning from soft targets does not by itself harm this network.

\section{Discussion}

Relearning helps every family although the memory is no more accurate than the student it is built from, which suggests that review passes on the relations among classes and a per-image confidence \cite{hinton2015distilling,furlanello2018born} rather than better answers, and, for the AdamW families, a restraint on fitting the labels more sharply than the memory does late in each episode. For ResNet-18, whose student falls far below its memory after every reheat, review may instead repair what the reheat erased. Review and practice also divide the work. Both keep the likelihood from drifting across episodes, while practice on dreams supplies most of the robustness that model annealing adds over snapshot distillation, beyond the share that soft targets alone already bring.

For ResNet-18 a reheat does not so much open a new study session as undo much of the last one. Several properties set this network apart: it alone is trained with SGD, without gradient clipping or stochastic depth, it alone uses batch normalization, and its learning rate is the largest value of its sweep. ConvNeXt, convolutional but trained with AdamW, keeps its fit through reheats, so convolution is not what makes the difference. The sweep chose each peak for a network at the start of training, yet a reheat applies it to a network that has just cooled, and warm restarts and snapshot ensembles that reheated SGD-trained residual networks with batch normalization to lower peaks reported gains \cite{loshchilov2017sgdr,huang2017snapshot}. This points to the peak rate, and to peaks matched to the cooled network as the remedy.

Nothing was tuned on Tiny-ImageNet or Imagewoof, and the gain of the AdamW families carries over to both: +1.7 points over cross-entropy on Tiny-ImageNet (8 of 9 configurations) and +0.5 on Imagewoof (8 of 12), against +1.8 on CIFAR-100, where the form of the memory was chosen. Because the memory speaks only through predictions, review could also draw on models whose weights cannot be shared, as in query-based collaborative learning \cite{ruzbahani2026dtqcl}. Fine-tuning pretrained networks, a common and strong choice when labels are few \cite{kolesnikov2020bit}, is a natural next setting for review and consolidation.

\section{Conclusion}

Model annealing turns the memorization phase of small-data training into a cycle of review, practice and consolidation. For the families trained with AdamW it raises accuracy, likelihood and robustness together, and it matches the strongest self-distillation baseline in accuracy while surpassing it in calibration and robustness. Review needs something to build on. Where a reheat erases most of what a network had learned, as it does for ResNet-18 under SGD, the evidence favors longer episodes or lower peaks. With little data, reviewing the whole notebook and practicing on dreams keep the likelihood of the student stable from one session to the next. Episode lengths and reheat peaks matched to each network, and memories that weight each snapshot by its validation likelihood, are natural next steps.

\section*{Acknowledgment}

This research was funded by MetaKet Inc. AI tools were used to polish the text, correct grammar, improve readability, and debug code.

\bibliographystyle{IEEEtran}
\bibliography{refs}

\end{document}